\documentclass{article}

\ifdefined\ANON
  \usepackage[dblblindworkshop]{neurips_2026}
  \workshoptitle{Machine Learning for Systems}
\else
  \usepackage[preprint]{neurips_2026}
\fi

\usepackage[utf8]{inputenc}
\usepackage[T1]{fontenc}
\usepackage{hyperref}
\usepackage{url}
\usepackage{booktabs}
\usepackage{amsfonts}
\usepackage{nicefrac}
\usepackage{microtype}
\usepackage{xcolor}
\usepackage{graphicx}
\usepackage{amsmath}
\usepackage{amssymb}
\usepackage{array}
\usepackage{enumitem}
\setlist{nosep,leftmargin=*}
\usepackage{titlesec}
\titlespacing*{\section}{0pt}{2.0ex plus 0.5ex minus .2ex}{1.0ex plus .1ex}
\titlespacing*{\subsection}{0pt}{1.6ex plus 0.4ex minus .2ex}{0.6ex plus .1ex}
\titlespacing*{\paragraph}{0pt}{1.4ex plus 0.3ex minus .2ex}{0.6em}

\title{GraphPass: A Study of Graph Neural Networks\\
       for Compiler Pass Selection}

\author{%
  Mohammed Saqlain\\
  Independent Researcher\\
  \texttt{saqlain.contact@gmail.com}
}

\begin{document}

\maketitle

\begin{abstract}
Compiler pass selection is usually governed by fixed pipelines
(\texttt{-O2}, \texttt{-O3}) or hand-written heuristics.
We study whether a graph neural network (GNN) over a program's
control-flow graph (CFG) can select among a fixed menu of five pass
sequences in a single forward pass.
\textbf{GraphPass} encodes each CFG with a graph attention (GAT) or
graph convolutional (GCN) encoder and classifies it into one of
the five sequences.
On a synthetic benchmark of 5{,}000 generated CFGs in five
program families, both encoders agree with the benchmark's reference
label on about 92\% of held-out programs, versus 68.4\% for a
hand-written rule-based heuristic and 85\% for a flat MLP.
An edge ablation indicates that this gain comes from per-block
processing rather than from CFG edges: a GCN given the same
basic-block features with all edges removed scores 95.0\%, versus 92.2\% with the true CFG.
In this benchmark the label is determined by the program
family, so the task amounts to family recognition.
Real-compiler speedup is not measured.
Inference costs $O(|V|+|E|)$ per CFG, measured at 1.7\,ms for a typical
program and about 0.8\,s for a 100{,}000-block graph on one CPU core.
\end{abstract}

\section{Introduction}
\label{sec:intro}

Modern optimizing compilers such as LLVM~\cite{lattner2004llvm}
expose dozens of analysis and transformation passes that can be
applied in many different orders.
The standard approach, a fixed sequence such as \texttt{-O2} or
\texttt{-O3}, applies the same ordering to every program, although
the best ordering is program dependent~\cite{fursin2008milepost}.
Iterative compilation~\cite{chen2012iterative} searches for good
orderings by repeatedly compiling and running a program, which is
expensive, and existing machine learning approaches~\cite{ashouri2018survey}
often summarize a program as a flat feature vector.
Graph representations of programs~\cite{cummins2021programl,brauckmann2020compiler}
and graph-based pass-ordering methods~\cite{liang2023coreset} instead keep relational structure.

A natural hypothesis is that the control-flow graph (CFG) carries
information about which passes help, and that GNNs~\cite{kipf2017gcn,velickovic2018gat}
can use it.
This paper tests that hypothesis on a synthetic benchmark.
\emph{On this benchmark}, most of the accuracy comes from processing
basic-block features rather than from the edges between them.

\textbf{Contributions.}
\begin{enumerate}
  \item We cast pass selection as $K$-way graph classification over CFGs
    with an explicit, fixed output space of $K{=}5$ pass sequences
    (Section~\ref{sec:method}), and give its inference cost as
    $O(|V|+|E|)$, measured empirically (Section~\ref{sec:latency}).
  \item We present GraphPass with two interchangeable encoders, a GAT
    (GraphPass-GAT) and a GCN (GraphPass-GCN), which reach statistically
    indistinguishable accuracy.
  \item We run an edge ablation (true edges vs.\ no edges vs.\ random
    edges, identical node features) and a size-generalization test.
  \item We characterize the benchmark: labels are
    a deterministic function of the generator's program family, so
    accuracy measures agreement with a synthetic reference rule.
\end{enumerate}

\section{Background and Related Work}
\label{sec:related}

\paragraph{Compiler pass ordering.}
MILEPOST GCC~\cite{fursin2008milepost} used machine-learned models over
features extracted from the compiler's internal representation.
Surveys of ML-based autotuning~\cite{ashouri2018survey} cover flag
selection and phase ordering at length.
Deep learning has been used to learn optimization heuristics directly
from source~\cite{cummins2017end}, and reinforcement learning has been used
for phase ordering in high-level synthesis, for example AutoPhase~\cite{haj2020autophase}.
Closest to our setting, Liang et al.~\cite{liang2023coreset} learn
LLVM pass orders with a coreset-based dataset and normalized value
prediction.
The TpuGraphs dataset~\cite{phothilimthana2023tpugraphs} provides graph-level
performance-prediction targets for tensor computation graphs.
GraphPass targets a small, fixed menu of pass sequences on synthetic CFGs.

\paragraph{GNNs for program analysis.}
ProGraML~\cite{cummins2021programl} introduced a graph representation
of programs for data-flow-style compiler tasks.
Brauckmann et al.~\cite{brauckmann2020compiler} evaluated compiler-derived
graph representations (ASTs and control/data-flow graphs) with GNNs for heterogeneous device
mapping and thread coarsening.
Our setting differs in that the target is a pass-sequence choice,
and in that we explicitly test whether the edges, rather than the node features,
carry the signal.

\section{Method: GraphPass}
\label{sec:method}
\begin{figure}[t]
\centering
\includegraphics[width=\textwidth]{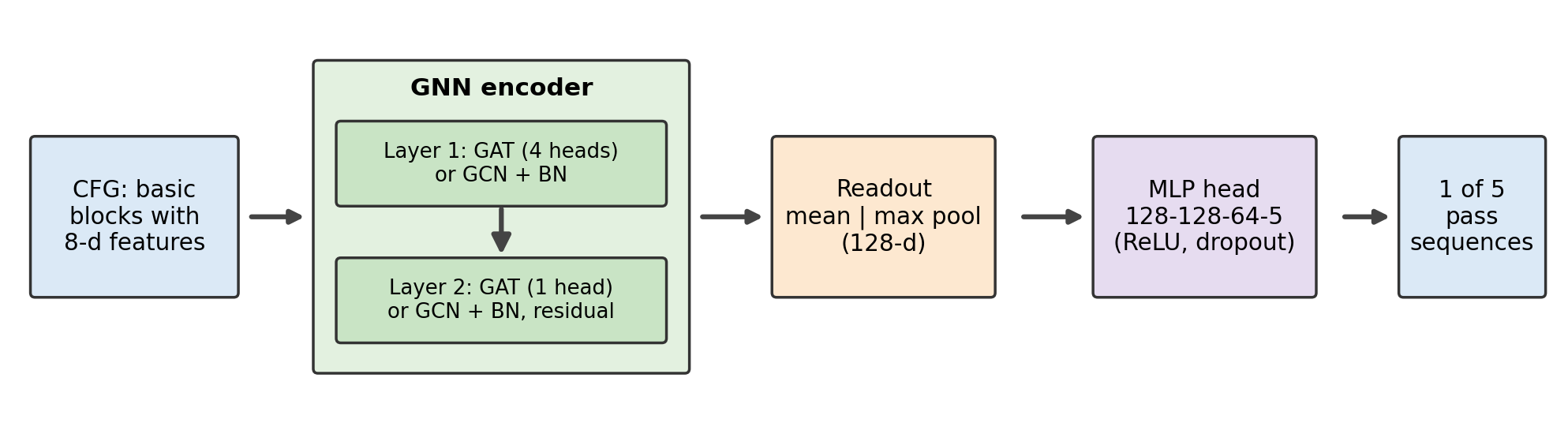}
\caption{GraphPass pipeline (GAT variant shown; the GCN variant replaces
  the two GAT layers with two GCN layers and uses a 128--128--5 head).
  A CFG whose nodes are basic blocks with 8-dimensional features is
  encoded by two graph layers, pooled by concatenating mean and max
  readouts, and classified by an MLP into one of five pass sequences.}
\label{fig:arch}
\end{figure}

\subsection{Output Space}
\label{sec:output}
The model does not generate arbitrary pass orderings.
We define ten LLVM-style passes (\texttt{mem2reg}, \texttt{instcombine},
\texttt{simplifycfg}, \texttt{loop-unroll}, \texttt{gvn}, \texttt{licm},
\texttt{dce}, \texttt{inline}, \texttt{sccp}, \texttt{reassociate}) and
hand-design $K{=}5$ fixed sequences over them (Table~\ref{tab:seqs}).
The model is a $K$-way graph classifier with a softmax over these five
classes, so its output is always exactly one non-empty sequence (each
has at least four passes) and chance accuracy is 20\%.
Sequences are \emph{not} compiled or executed in this study.

\begin{table}[t]
\centering
\caption{The five candidate pass sequences (output classes) and the
  program family whose reference label they are.}
\label{tab:seqs}
\small
\begin{tabular}{llp{6.2cm}}
\toprule
\textbf{Class} & \textbf{Family} & \textbf{Pass sequence} \\
\midrule
Light        & simple        & mem2reg, instcombine, simplifycfg, dce \\
Inline-First & call-heavy    & mem2reg, inline, instcombine, simplifycfg, dce \\
ValueNum     & branch-heavy  & mem2reg, gvn, sccp, instcombine, simplifycfg, dce \\
LoopOpt      & loop-heavy    & mem2reg, licm, loop-unroll, instcombine, simplifycfg, dce \\
Full         & compute-heavy & mem2reg, inline, gvn, licm, loop-unroll, instcombine, simplifycfg, dce \\
\bottomrule
\end{tabular}
\end{table}

\subsection{Program Graphs}
A program is a CFG $G=(V,E)$ whose nodes are basic blocks and whose
directed edges are possible control transfers.
Each node has $d=8$ features: a node-type code (entry, exit, loop header,
call site, branch, normal), normalized instruction count, $\phi$-node count,
a memory-operation flag, successor count, a loop-depth proxy, a call
count and an arithmetic-density flag.
\emph{Note that three of these (successor count, $\phi$ count, loop-header
indicator) are computed from the CFG edges}, so a model that never sees
edges still receives a local summary of topology; we return to this
in Section~\ref{sec:edges}.

\subsection{Encoders and Readout}
The GAT encoder~\cite{velickovic2018gat} updates node $v$ at each layer as
\begin{equation}
  \mathbf{h}_v' = \big\Vert_{k=1}^{K_h}\,\mathrm{ELU}\Big(
    \sum_{u \in \mathcal{N}(v)\cup\{v\}}
    \alpha^{k}_{uv}\, \mathbf{W}^{k} \mathbf{h}_u\Big),
  \label{eq:gat}
\end{equation}
with attention coefficients $\alpha^k_{uv}$ from a single-layer
feed-forward scorer, where $\Vert$ is head concatenation.
Layer~1 uses $K_h{=}4$ heads of width 64 (256 dimensions),
followed by batch normalization, ELU and dropout.
Layer~2 uses one head of width 64, batch normalization, and a
residual connection from a linear projection of the input features.
The readout concatenates mean and max pooling over nodes
($\mathbb{R}^{128}$), followed by an MLP $128\!\to\!128\!\to\!64\!\to\!5$
with ReLU and dropout 0.3.
The GAT model has 45{,}701 parameters.
GraphPass-GCN replaces the GAT layers by two 64-wide GCN
layers~\cite{kipf2017gcn} with batch normalization and ReLU, and uses a
128--128--5 head.

\subsection{Training}
We use cross-entropy loss and Adam (learning rate $3\!\times\!10^{-3}$,
weight decay $10^{-4}$), batch size 64, a cosine schedule, gradient-norm
clipping at 1.0 (GAT) and checkpoint selection by validation accuracy.
GraphPass-GAT trains for 80 epochs; for the GCN and MLP we report both
40 epochs and the same 80-epoch budget as the GAT.
The 5{,}000 graphs are split randomly 70/15/15 (3{,}500 / 750 / 750).
Classes are exactly balanced overall (1{,}000 each); the split is random,
not stratified.
Results are mean $\pm$ std over five training seeds unless stated.
Models are implemented in PyTorch Geometric~\cite{fey2019pyg}.

\section{Experiments}
\label{sec:experiments}

\subsection{Benchmark and Baselines}
\label{sec:data}
The benchmark consists of 5{,}000 procedurally generated CFGs, 1{,}000 in each of five
families (simple, loop-heavy, call-heavy, branch-heavy, compute-heavy),
with 8--49 basic blocks (mean 23.5) and on average 39.6 edges.
Family-specific parameters (block-count range, back-edge rate,
call rate, branch rate, arithmetic density) shape each graph; block-count ranges are
simple 8--19, loop-heavy 15--39, call-heavy 20--49, branch-heavy 10--29 and
compute-heavy 12--34.
\textbf{Labels are not obtained from a compiler cost model.}
Each family is assigned one fixed class (Table~\ref{tab:seqs}); the label is a
deterministic function of the family.
Accuracy therefore measures agreement with this reference mapping, and
the learning problem is, in effect, recognizing the generator's family from a graph.
We discuss consequences in Section~\ref{sec:discussion}.

Baselines:
\textbf{Random}, \textbf{Always-Full} (always the heaviest sequence,
a stand-in for \texttt{-O3}) and \textbf{Always-Light} (a stand-in for
\texttt{-O1});
a \textbf{rule-based heuristic}, a hand-written decision list over mean
loop-depth, call count, fraction of branch nodes and arithmetic density;
a \textbf{flat MLP} on the mean-pooled node features (no graph, no per-node
processing); and a \textbf{GCN} (GraphPass-GCN).
The heuristic was written on the same features with untuned thresholds
(Section~\ref{sec:discussion}).

\subsection{Main Results}
\begin{table}[t]
\centering
\caption{Test accuracy (750 programs). Mean $\pm$ std over 5 seeds;
  baselines without training are a single evaluation. ``Ep.'' is the
  number of training epochs. Chance is 20\%.}
\label{tab:main}
\begin{tabular}{lcc}
\toprule
\textbf{Method} & \textbf{Ep.} & \textbf{Accuracy (\%)} \\
\midrule
Random                      & --- & 19.9 \\
Always-Full ($\approx$\texttt{-O3})   & --- & 21.1 \\
Always-Light ($\approx$\texttt{-O1})  & --- & 20.8 \\
Rule-based heuristic        & --- & 68.4 \\
\midrule
Flat MLP (mean-pooled features) & 40 & $85.0 \pm 0.2$ \\
Flat MLP (mean-pooled features) & 80 & $85.3 \pm 0.4$ \\
GraphPass-GCN               & 40 & $91.8 \pm 0.7$ \\
GraphPass-GCN               & 80 & $92.4 \pm 0.4$ \\
GraphPass-GAT, 1 head       & 80 & $90.1 \pm 0.7$ \\
GraphPass-GAT, 4 heads      & 80 & $92.0 \pm 0.8$ \\
\bottomrule
\end{tabular}
\end{table}

Table~\ref{tab:main} shows both GraphPass encoders at about 92\%,
23.6 points above the heuristic (68.4\%) and about 7 points above the flat MLP.
These are \emph{accuracy} differences against a synthetic reference
label; speedup is not measured.
GCN and 4-head GAT are statistically indistinguishable at equal budget
($92.4\pm0.4$ vs.\ $92.0\pm0.8$); a single GAT head is about two points
lower.
Per-class accuracy (seed~0 GAT, 93.3\% overall):
loop-heavy 100\%, branch-heavy 99.3\%, compute-heavy 96.2\%, simple 94.2\% and
call-heavy 75.2\%, with most errors assigned to the Light class (30 of 137).
The heuristic reaches 93--99\% on simple, loop-heavy and compute-heavy
programs and 26.3\% and 18.8\% on call-heavy and branch-heavy ones,
where its fixed thresholds rarely fire.

\subsection{Do CFG Edges Matter? An Edge Ablation}
\label{sec:edges}
Comparing a GNN with a flat MLP confounds two factors: the GNN sees every
node's features, whereas the flat MLP sees only their mean.
We hold the node features fixed and vary
only the edges: the \emph{true} CFG, \emph{no} edges (each block sees only itself), or
\emph{random} edges with the same edge count but random endpoints.
Features are unchanged in all three conditions.

\begin{table}[t]
\centering
\caption{Edge ablation, identical node features. Test accuracy (\%),
  mean $\pm$ std over $n$ seeds (GCN: $n{=}5$, GAT: $n{=}3$).
  Values come from a separate run from Table~\ref{tab:main}
  (different random streams), so compare only within this table.}
\label{tab:edges}
\begin{tabular}{lccc}
\toprule
\textbf{Encoder} & \textbf{True CFG} & \textbf{No edges} & \textbf{Random edges} \\
\midrule
GraphPass-GCN & $92.2 \pm 0.4$ & $\mathbf{95.0 \pm 0.3}$ & $86.5 \pm 0.7$ \\
GraphPass-GAT & $\mathbf{93.1 \pm 0.4}$ & $87.0 \pm 1.3$ & $89.5 \pm 1.0$ \\
\bottomrule
\end{tabular}
\end{table}

Table~\ref{tab:edges} gives a mixed answer.
Corrupting the topology (random edges) hurts both encoders relative to the true CFG, so the models
do use edges when they are present.
But \emph{removing} edges entirely does not hurt the GCN, which improves to 95.0\%,
the best result in this paper, while the GAT loses six points without edges.
Whether true CFG edges add information beyond the blocks' own
features therefore depends on the encoder.
Two caveats apply: the successor-count, $\phi$-count and loop-header features are derived
from the edges, so a no-edge model still receives a local topology summary;
and the GAT and GCN differ in other respects (activation, residual, head depth).
A per-node GCN without edges is 9.7 points above the flat
MLP (85.3\%), so the gain of the GNNs over the flat baseline comes mostly from
\emph{processing individual blocks and max-pooling} rather than from graph relations.

\subsection{Program Size}
\label{sec:size}
Table~\ref{tab:size} bins the 750 test programs by block count (seed~0 GAT, 93.3\%).
Accuracy \emph{rises} with size rather than staying flat, and the heuristic
is non-monotone rather than steadily degrading.
Size is confounded with family (e.g.\ only call-heavy programs reach
49 blocks while simple programs stop at 19), so size and family effects are intertwined.
The benchmark has at most 49 blocks.

\begin{table}[t]
\centering
\caption{Test accuracy (\%) by number of basic blocks (seed~0).}
\label{tab:size}
\begin{tabular}{lcccc}
\toprule
\textbf{Blocks} & 8--14 & 15--21 & 22--29 & 30--49 \\
\textbf{\# programs} & 162 & 193 & 187 & 208 \\
\midrule
GraphPass-GAT & 84.0 & 91.2 & 96.3 & 100.0 \\
Heuristic     & 75.3 & 64.2 & 60.4 & 74.0 \\
\bottomrule
\end{tabular}
\end{table}

As a size-generalization test we train only on programs with at most 22 blocks
(1{,}791 graphs) and test on test programs with more than 22 blocks (374 graphs, no simple programs).
GAT reaches $98.0\pm0.1$\% (3 seeds), GCN $97.6\pm0.5$\%, flat MLP $94.6\pm0.7$\% and the
heuristic 68.2\%.
This subset is dominated by the families that are easiest to separate.

\subsection{Inference Cost}
\label{sec:latency}
For a fixed model, inference is linear in graph size:
each GNN layer costs $O(|V|\,d\,d' + |E|\,d')$ per head, so the
forward pass is $O(|V|+|E|)$.
Constructing the CFG is also at least $O(|V|+|E|)$.
It is constant only in the number of candidate sequences and in the number of
compilations, in contrast to iterative compilation, which needs on the order of $K$ compile-and-run cycles.
We measured the GAT forward pass (batch size one, one CPU thread) in Table~\ref{tab:latency}.
Programs from our benchmark take 1.7\,ms (median; p95 2.0\,ms).
Synthetic chain-like graphs with average degree two show the expected
growth: latency is about 8.3\,$\mu$s per block at 100{,}000 blocks.
Extrapolating linearly, a graph with a million blocks would take on the order of
8\,s on this core (an extrapolation, not a measurement).
CFG-extraction time from real LLVM-IR and the fraction of real compile time are not measured.

\begin{table}[t]
\centering
\caption{GAT forward-pass latency vs.\ graph size (median, one thread).}
\label{tab:latency}
\begin{tabular}{lccccc}
\toprule
\textbf{Blocks} & 25 & 100 & 1{,}000 & 10{,}000 & 100{,}000 \\
\midrule
\textbf{Latency (ms)} & 1.6 & 2.1 & 7.1 & 64.7 & 827 \\
\bottomrule
\end{tabular}
\end{table}

\subsection{Learned Representations}
\label{sec:tsne}
\begin{figure}[t]
\centering
\includegraphics[width=0.62\textwidth]{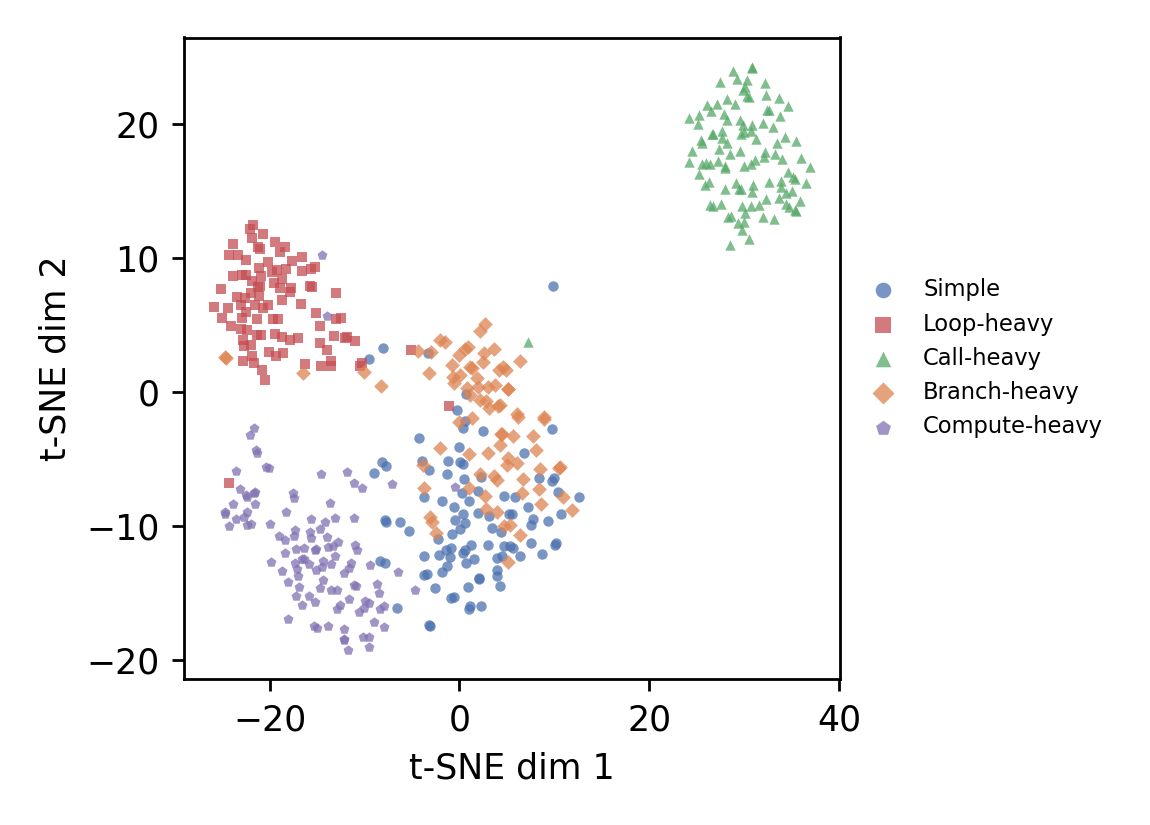}
\caption{t-SNE~\cite{maaten2008tsne} of GraphPass-GAT graph embeddings
  (500 test programs, seed~0), colored by program family.
  Silhouette score by family: 0.46 in the 2-D projection, 0.20 in the
  original embedding space.}
\label{fig:tsne}
\end{figure}
Figure~\ref{fig:tsne} shows embeddings that cluster by family.
Clustering by family is expected, since the training label is in one-to-one
correspondence with the family.
The moderate silhouette (0.20 in the embedding space) agrees with
the confusion between call-heavy and simple programs.

\section{Discussion and Limitations}
\label{sec:discussion}

On this benchmark, GNN encoders classify the generator's family
more accurately than a hand-written heuristic and a flat feature baseline, and inference is cheap
for function-sized CFGs.
A GAT and a GCN perform equally well, so we name both as GraphPass variants.

\paragraph{Scope of the results.}
(i)~The edge ablation attributes the accuracy mainly to per-block
processing (Section~\ref{sec:edges}).
(ii)~Because labels are a function of the family, the model may be learning
family-specific feature statistics; separating these from factors that govern
optimization outcomes requires labels from real compilation.
(iii)~The natural evaluation metric is the speedup (or code size) obtained by running the
chosen sequence, ideally with regret for wrong choices rather than 0/1 accuracy; neither is measured here.
(iv)~The heuristic baseline is hand-written, so the margin over it is not a margin over a tuned
production heuristic or over \texttt{-O3}.
The ``Always-Full'' row is a label-agreement baseline, not measured \texttt{-O3} performance.

\paragraph{Limitations.}
All data are synthetic, with only 5 candidate sequences, 8--49 blocks, a single
CPU, function-level CFGs, no real LLVM-IR parsing, and no interprocedural information.
GAT ablations use three seeds.
Real LLVM-IR inputs, labels obtained by compiling and timing the candidate sequences on
benchmark suites, and larger action spaces are outside the scope of this study.

\section{Conclusion}
\label{sec:conclusion}

We built GraphPass, a GNN pass selector over CFGs with an explicit five-sequence output space, and
evaluated it on a synthetic benchmark.
Both GAT and GCN encoders reach about 92\% agreement with the reference label and inference is $O(|V|+|E|)$ at
1.7\,ms per typical program.
An edge ablation attributes the gain mainly to per-block processing rather than to CFG edges,
and the benchmark labels are determined by program family.
Labels from measured speedups on real programs are outside the scope of this study.

\bibliographystyle{unsrt}

\end{document}